\documentclass[letterpaper]{article} % DO NOT CHANGE THIS
\usepackage{aaai2027}  % DO NOT CHANGE THIS
\usepackage[hyphens]{url}  % DO NOT CHANGE THIS
\usepackage{graphicx} % DO NOT CHANGE THIS
\usepackage{array} 
\usepackage{multirow} 
\usepackage{amsmath} 
\usepackage{amssymb}
\usepackage[most]{tcolorbox}
\usepackage{natbib}  % DO NOT CHANGE THIS AND DO NOT ADD ANY OPTIONS TO IT
\usepackage{caption} % DO NOT CHANGE THIS AND DO NOT ADD ANY OPTIONS TO IT
\usepackage{algorithm}
\usepackage{algorithmic}
\usepackage[hidelinks]{hyperref}
\usepackage{newfloat}
\usepackage{listings}
\DeclareCaptionStyle{ruled}{labelfont=normalfont,labelsep=colon,strut=off} % DO NOT CHANGE THIS
\floatstyle{ruled}
\newfloat{listing}{tb}{lst}{}
\floatname{listing}{Listing}

\usepackage{booktabs}

\usepackage{xspace}

\makeatletter
\DeclareRobustCommand\onedot{\futurelet\@let@token\@onedot}
\def\@onedot{\ifx\@let@token.\else.\null\fi\xspace}

\makeatother

\title{\Large SketchDirector: Complex Scene Sketching via Agentic Layout Reasoning and Detail Refinement}
\author{
    Yiqi Liang and Xiaodong Cun\corresponding
}
\affiliations{
    GVC Lab, Great Bay University\\
    \href{https://yiki77.github.io/SketchDirector/}
    {https://yiki77.github.io/SketchDirector/}
}

\nocopyright

\makeatletter
\let\tempotmd@orig@maketitle\@maketitle
\def\@maketitle{%
  \tempotmd@orig@maketitle
  \begin{center}
    \makebox[\textwidth][c]{%
      \includegraphics[width=1.00\textwidth]{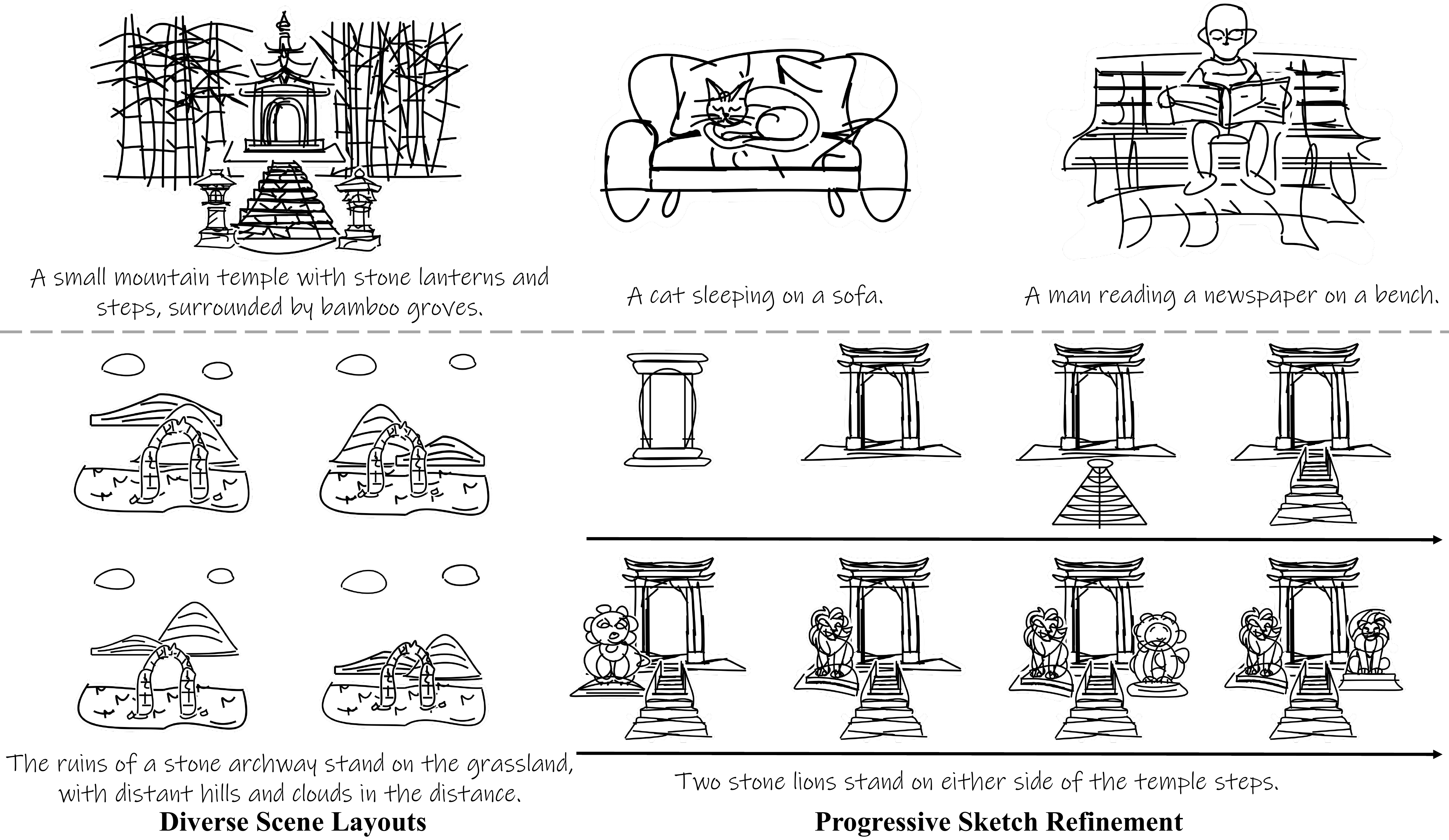}%
    }
    \vspace{-2em}

    {%
      \captionsetup{hypcap=false}
      \captionof{figure}{\textbf{Visual results of SketchDirector}. The top row shows scene sketches depicting diverse concepts. The bottom-left row shows rich layout variations for the same prompt, highlighting the effectiveness of agentic reasoning for automatic scene composition. The bottom-right row shows the progressive sketch refinement process, which enhances sketch details while preserving sketch structure and spatial layout.}
      \label{fig:teaser}
    }%
  \end{center}
}
\makeatother

\begin{document}

\maketitle
\begin{abstract}
% Scene sketch has great potential in animation, proof of concept, and interactive design. 
Existing sketch generation methods often rely on image guidance and are largely limited to individual objects or simplified scenes, leaving text-guided complex scene sketching underexplored. 
We present SketchDirector, an agentic framework that mimics the human sketching workflow for automatic multi-concept scene sketching.
Given a scene-level text prompt, SketchDirector first employs a large language model~(LLM) to decompose the scene into multiple draft assets, where each is represented by a set of B\'ezier curves. Then, an agentic reasoning loop iteratively plans and executes their spatial arrangement, guided by self-evolving memory. Finally, a detail refinement module leverages diffusion priors to enhance sketch quality while preserving geometric structure. To support future research and applications in multi-concept scene sketch generation, we further introduce the \textit{MCSketch8K} dataset, comprising editable vector assets, compositional layouts, and object-level manipulation annotations. Extensive experiments show that SketchDirector outperforms state-of-the-art baselines in semantic and compositional quality while effectively preserving overall contour organization during refinement. The project page is: \href{https://yiki77.github.io/SketchDirector/}
    {https://yiki77.github.io/SketchDirector/}.
% Specifically, we first use a Large Language Model~(LLM) to decompose the whole scene prompt into editable object units. Then, an agentic layout reasoning loop iteratively plans, executes, and verifies the arrangements through multi-modal feedback and hierarchical self-evolving. Then, a structure vector refinement is utilized to further refine the generated sketch from the diffusion prior.  Finally, to facilitate future research and application on multi-concept scene sketch generation, we introduce a multi-concept scene sketch dataset with editable vector assets, compositional layouts, and manipulation annotations.
% Extensive experiments demonstrate that SketchDirector outperforms state-of-the-art baselines in object modularity, layout alignment, and structure-preserving refinement.
\end{abstract}
  
\section{Introduction}
\begin{figure}[t]
    \centering
    \includegraphics[width=1.0\linewidth]{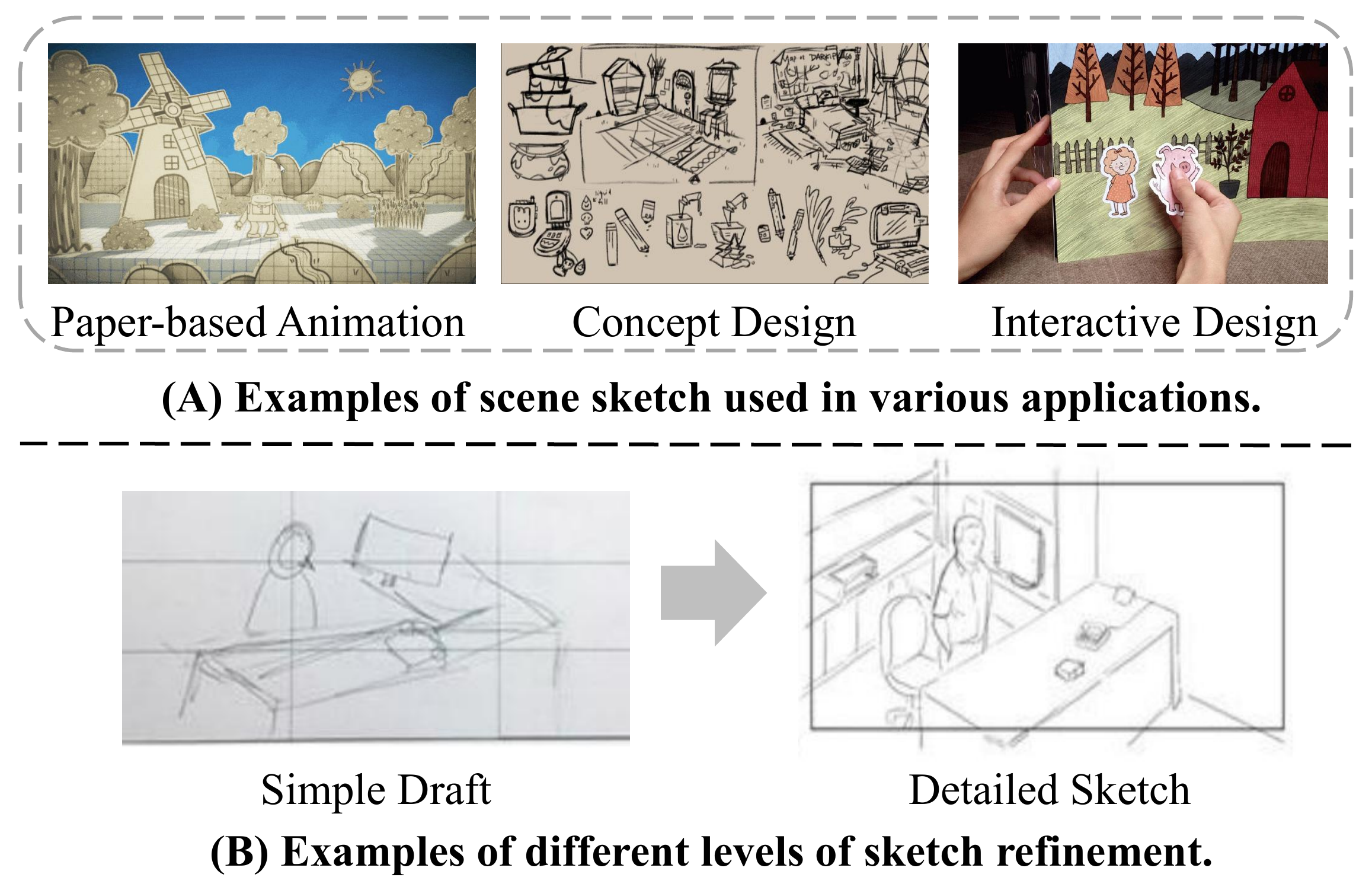} % Reduce the figure size so that it is slightly narrower than the column.
    \vspace{-2em}
    \caption{\textbf{Motivation of multi-concept scene sketch generation.} (A) Widely used in animation, concept art, and design. (B) Artists adjust sketch refinement from draft to detailed sketch preserving structure for different goals.}
    \label{f1-application}
    \vspace{-2em}
\end{figure}

Sketching is a concise visual representation that effectively captures the intentions of human beings.
As its scope expands from individual objects to multiple concepts, scene sketching has shown great potential in applications such as animation, proof of concept, and interactive design, as in Fig.~\ref{f1-application}~(A).
However, complex scene sketching requires more than depicting individual objects. It involves organizing multiple concepts, reasoning about their spatial relationships, and maintaining coherent details within a unified scene. 
Therefore, generating semantically aligned scene sketches with reasonable layouts and coherent details remains a big challenge.

Previous automatic scene sketching generation methods~\cite{li2019photo, li2019im2pencil, chan2022learning, vinker2023clipascene, liang2025mross} primarily operate at the holistic scene level, treating scene sketches as an indivisible entity without object representations, thereby hindering local editing and object reuse.
Recent studies~\cite{wu2025sketchtriplet, zhang6138582sketchany, zhan2025coprosketch} have explored object-aware generation, but objects are often treated as predefined components or auxiliary structures.
Consequently, these methods lack dynamic control, layout planning, and iterative optimization for individual objects, which limits their potential as visual components.

% 这一段主要讲我们根据之前的一些方法和现实中的一些 creation，得到的观察。引出下一段介绍我们的 agentic 的方法。
% In contrast, artists in real-world creative practice adapt the level of sketch refinement to their creative goals, producing sketches that range from rough drafts to highly detailed renderings.~(as Fig.~\ref{f1-application}~(B)). \xd{On the other hand, sketches need to be iteratively edited, with grouping and rearrangement for better presentation. As for the vector sketching in the digital space, it is also naturally represented by a set of B\'ezier curves and can be easily modified by code.}
In contrast, real-world sketch creation is inherently iterative: artists continually revise, group, and rearrange individual elements to improve the overall composition and presentation. They also adapt the level of refinement to suit their creative goals, producing sketches that range from rough drafts to highly detailed renderings, as illustrated in Fig.~\ref{f1-application}~(B). Digital vector sketches are particularly well suited to such flexible and iterative workflows because they are naturally represented as collections of B\'ezier curves, enabling efficient programmatic manipulation.

% Some methods~\cite{vinker2022clipasso,vinker2023clipascene} achieve refinement control by adjusting stroke counts under image guidance. However, how to flexibly refine text-guided scene sketches while preserving their underlying structures remains underexplored.

% These phenomena inspired us to apply the power of agentic coding agents~\cite{vinker2025sketchagent, du2026teach, liu2025multi, liang2025multi} to generate complex scene sketches.
% Inspired by the powerful capabilities of the agentic paradigm~\cite{vinker2025sketchagent, du2026teach, liu2025multi, liang2025multi}, 
We thus propose \textit{SketchDirector}, an automatic agentic framework~\cite{vinker2025sketchagent, du2026teach, liu2025multi, liang2025multi} for text-guided multi-concept scene sketch generation by mimicking the pipeline of human creators.
% Unlike existing agent-based approaches that mainly focus on stroke-level or part-level generation of objects~\cite{vinker2025sketchagent, du2026teach}, we extend agentic reasoning to generate a scene sketch.
% \xd{
Specifically,
% }
given a textual description, \textit{SketchDirector} first employs a Large Language Model~(LLM) based method to decompose the scene and generate each individual object as draft asset, each represented as a set of B\'ezier curves. 
To enable layout control between multiple asset, we introduce an operating action library comprising a set of executable operations, \emph{e.g.,} \texttt{move()}, \texttt{scale()}, to manipulate object attributes, enabling flexible adjustment of spatial relationships among objects.
Building upon this library, we propose an agentic reasoning loop, which formulates layout manipulation as an iterative operations and execution process.
To achieve reliable closed-loop reasoning, rather than predicting the entire scene layout at once, \textit{SketchDirector} progressively organizes draft assets through an iterative \emph{Plan--Execute--Verify--Correct} process.
At each step, \textit{SketchDirector} generates and executes object-level operations, updates the scene representation, and evaluates the resulting composition through geometric checks and multimodal verification.
When the layout fails verification, \textit{SketchDirector} revises its decisions and corrects the layout using a hierarchical self-evolving memory that leverages short-term failure reflections and accumulated long-term experiences.

After that, we introduce an additional detail refinement step to address the challenges of structural consistency in scene sketch distillation from the diffusion prior. 
Unlike traditional methods that treat B\'ezier curves as unstructured point sets, we treat object curve segments as the optimization primitives.
Then, by jointly optimizing local objects and the global canvas using pre-scene distillation, our method is able to generate scene sketches with varying levels of object refinement while maintaining consistency with the textual description. Finally, with the lack of available object-aware and editable scene sketch datasets, we construct a multi-concept scene sketch dataset called \textit{MCSketch8K} to facilitate future research and applications in scene sketch generation. 
The dataset comprises 1,000 scene sketches with over 4,000 disentangled vector object assets across 800 object categories. 
Each scene is annotated with compositional layouts and object-level manipulation operations, providing structured supervision for object decomposition, spatial arrangement reasoning, and controllable scene editing.

Our key contributions are summarized as:
\begin{itemize}
\item We propose \textit{SketchDirector}, an automatic framework for text-guided complex scene sketching with flexible object-level control and structure-preserving detail refinement.
\item \textit{SketchDirector} contains two key techniques to assist in scene sketch generation: \textit{i) agentic reasoning}, which formulates generation as a closed-loop process; \textit{ii) detail refinement}, which performs B\'ezier curve parametrization and progressively joint local-global vector optimization to refine scene sketches.
\item We construct \textit{MCSketch8K}, a multi-concept scene sketch dataset comprising vector object assets, compositional scene layouts, and object-level manipulation annotations.
\end{itemize}

\section{Related Work}

\noindent\textbf{Scene Sketch Generation.}
Scene sketch generation aims to translate visual or textual descriptions into coherent sketches. 
Early methods formulated this task as image-to-sketch translation~\cite{li2019photo, li2019im2pencil, chan2022learning}, converting photographs into rasterized contour drawings via supervised learning or adversarial training. 
Later studies increasingly adopt vector representations for scalable and editable sketch generation.
Image-conditioned methods such as Clipascene~\cite{vinker2023clipascene} and MROSS~\cite{liang2025mross} produce vector scene sketches from a reference image, while text-guided methods~\cite{frans2022clipdraw, jain2023vectorfusion, xing2023diffsketcher, xing2024svgdreamer, xing2025svgdreamer++} further enable open-domain vector sketch generation from textual descriptions. 
In particular, VectorFusion~\cite{jain2023vectorfusion} distills text-to-image diffusion priors into SVGs through score distillation, DiffSketcher~\cite{xing2023diffsketcher} directly optimizes B\'ezier curves with diffusion-based supervision, and SVGDreamer~\cite{xing2024svgdreamer} improves text-to-SVG generation through semantic-driven vectorization and particle-based score distillation.
Despite these advances, existing methods optimize the scene holistically, leaving objects implicitly encoded and limiting editing to strokes rather than objects.

\noindent\textbf{Multi-concept Sketch Generation.}
Recent studies explore structured scene generation by introducing different forms of concept decomposition and controllability.
For instance, SketchTriplet~\cite{wu2025sketchtriplet} assembles individual objects to construct scene datasets, although the objects mainly serve as fixed composition units without supporting layout planning or per-concept editing.
SketchAny~\cite{zhang6138582sketchany} separates foreground and background to improve structural coherence, yet provides limited control over individual concept instances.
CoProSketch~\cite{zhan2025coprosketch} represents objects with unsigned distance fields to enable progressive scene generation and improved editability, while generation is still performed at the scene level rather than through independent concept generation. 
Collectively, these methods highlight the value of concept-aware representations, yet objects remain passive elements rather than autonomous units for planning, interaction, and iterative refinement.

\noindent\textbf{Agents for Sketch Generation.}
The rapid development of multimodal large language models (MLLMs)  has introduced new opportunities for reasoning, planning, and iterative interaction into sketch generation.
For instance, SketchAgent~\cite{vinker2025sketchagent} introduces a language-driven sequential sketch generation framework, where an LLM-based agent can generate sequences of vector sketches and supports conversational sketch refinement. 
Building upon this, Du et al.~\cite{du2026teach} further explores controllable vector sketch generation by training a multimodal agent with part-level annotations and process-level feedback, enabling interpretable and locally editable sketch synthesis. 
In addition, AutoSketch~\cite{chin2025autosketch} investigates style-aware vector sketch completion, leveraging VLM-based iterative adjustment to preserve the abstraction level and stroke characteristics of input sketches during completion. Despite promising results, existing methods lack high-level scene reasoning and multi-object coordination. SketchDirector addresses this with an agentic framework for hierarchical planning, object-level control, and iterative refinement.

\begin{figure*}[t]
    \centering
\includegraphics[width=1.0\linewidth]{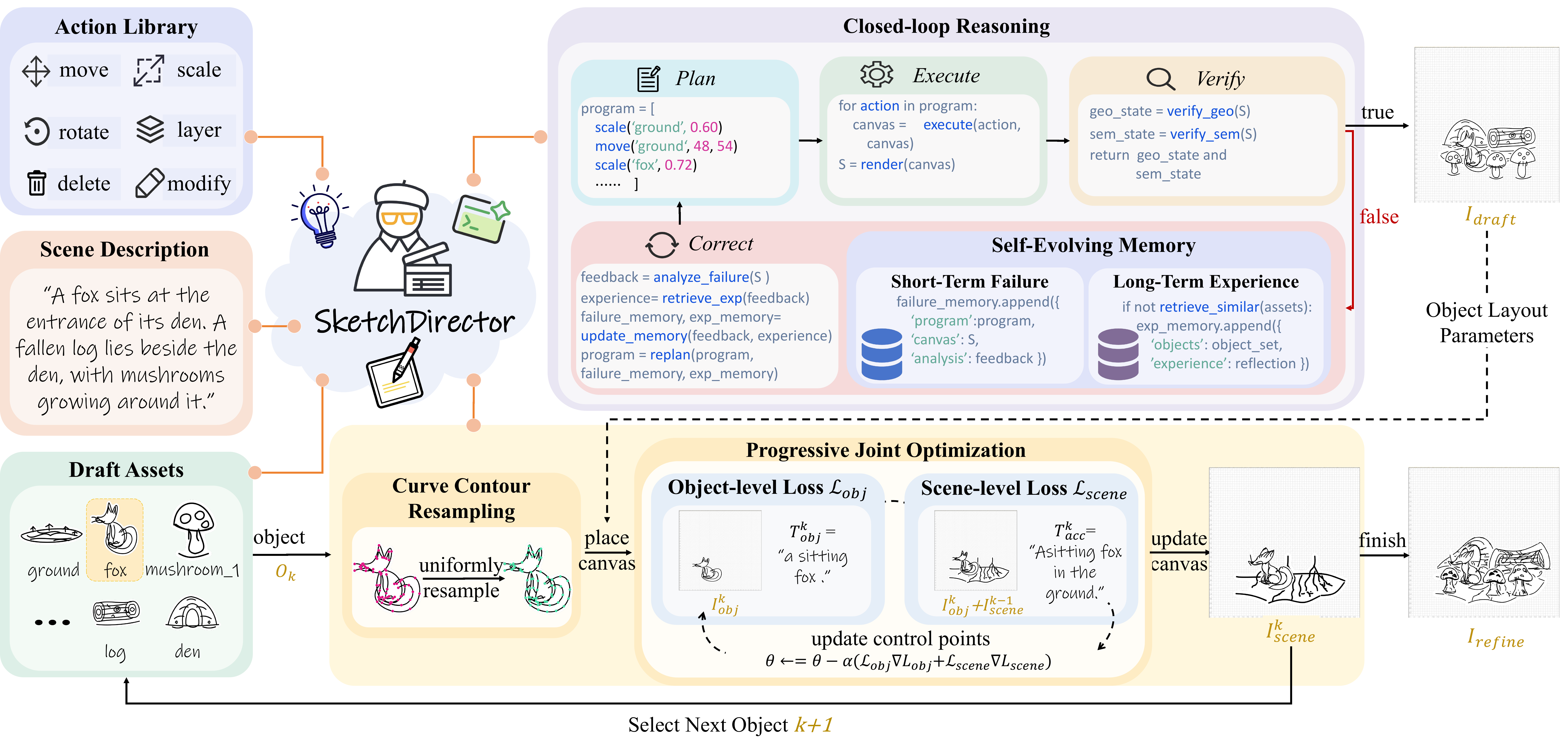} % Reduce the figure size so that it is slightly narrower than the column.
\vspace{-2em}
\caption{\textbf{\textit{SketchDirector} pipeline}. Given a scene description, \textit{SketchDirector} first generates object-level draft assets. The agent then progressively composes the scene through executable operations from the action library, maintaining scene programs and spatial asset attributes. A closed-loop process iteratively adjusts the layout, supported by self-evolving memory that incorporates short-term failure feedback and long-term experiences. Finally, detail refinement employs curve contour resampling and progressive joint optimization to enhance sketch details while preserving the established scene layout and object structures.}
\vspace{-1em}
\label{fig:main}
\end{figure*}
\begin{figure}[t]
    \centering
\includegraphics[width=1.0\linewidth]{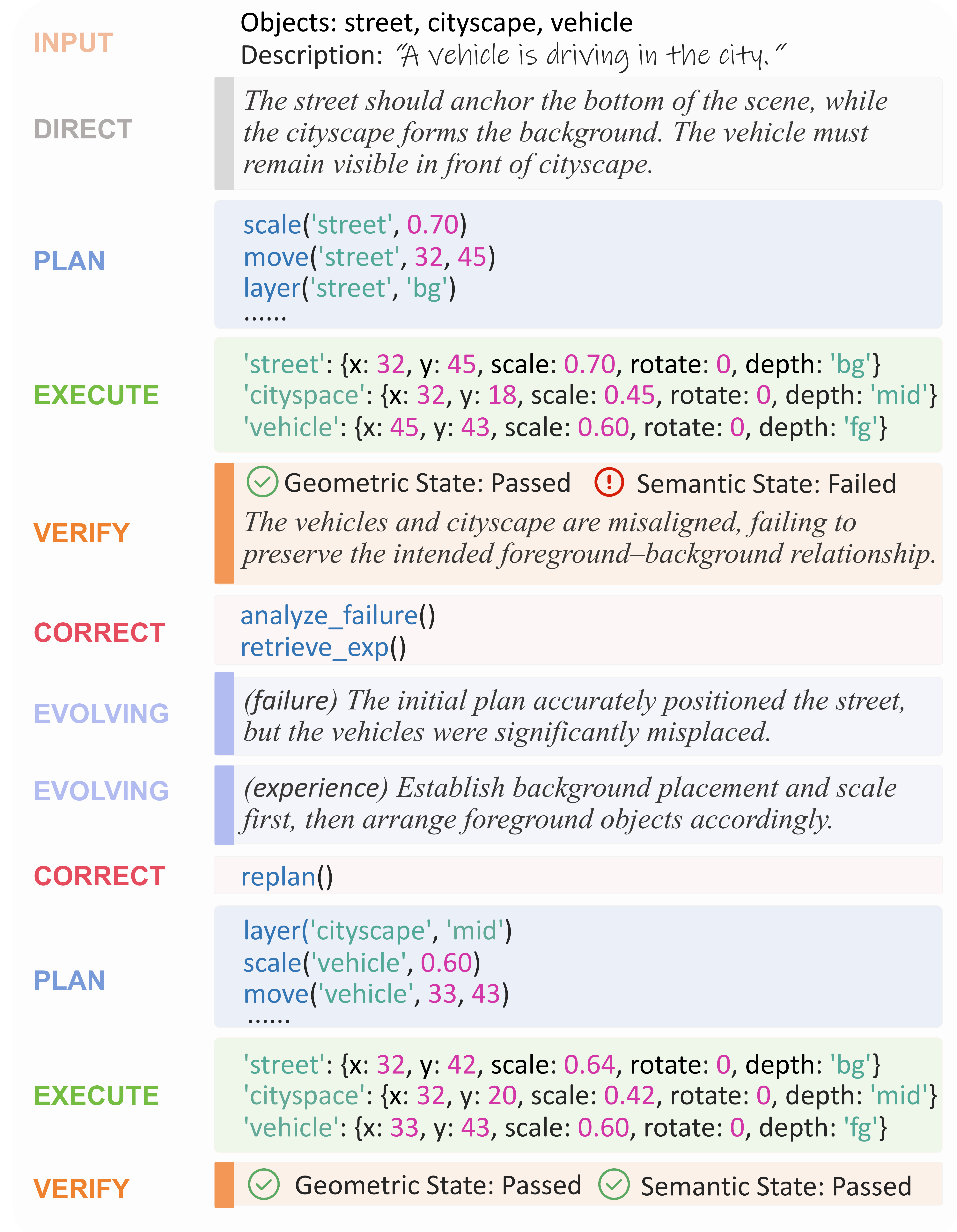} % Reduce the figure size so that it is slightly narrower than the column.
\vspace{-2em}
\caption{\textbf{Example agentic reasoning trajectory in \textit{SketchDirector}.}
A real log  illustrating how agents interleave layout planning, scene state updates, geometric-semantic verification, memory evolution, and correction.}
\vspace{-1em}
\label{fig:pipeline}
\end{figure}

\begin{figure}[t]
    \centering
\includegraphics[width=1.0\linewidth]{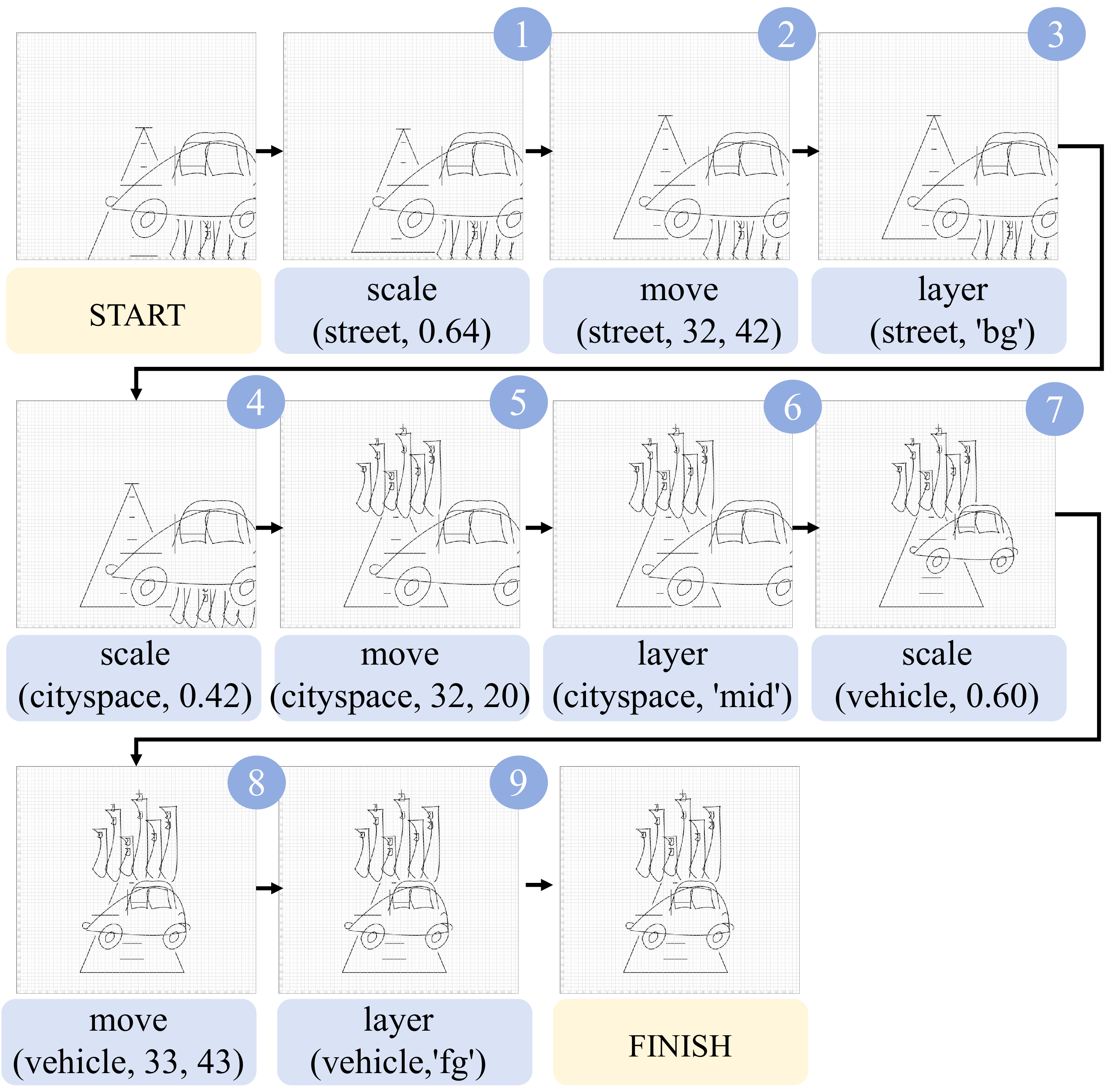} % Reduce the figure size so that it is slightly narrower than the column.
\vspace{-2em}
\caption{\textbf{A success example of planning and execution in \textit{SketchDirector}.}
Each operation updates both the executable program and the represented scene state, yielding a traceable trajectory from the initial drafts to the final layout.}
\vspace{-1em}
\label{fig:reason}
\end{figure}

\section{SketchDirector}
% \subsection{Overall Framework}
The goal of \textit{SketchDirector} is to generate editable multi-concept scene sketches from textual descriptions.
Each sketch is represented as a collection of SVG assets parameterized by black B\'ezier curves, enabling object-level editability and structural preservation.
Given a scene description, we first decompose it into individual concepts and generate corresponding vector draft assets~(Sec.~\ref{sec:step1}).
To support executable scene composition, we define an operating action library consisting of primitive operations that manipulate object attributes and update the scene state~(Sec.~\ref{sec:step2}).
Building upon these, an agentic layout reasoning step progressively composes the scene through a closed loop, where agents translate spatial intentions into executable editing programs and iteratively plan, execute, verify, and correct the scene composition~(Sec.~\ref{sec:step3}).
Finally, a detail refinement step further enhances sketch details through B\'ezier curve contour resampling and progressive joint optimization, while preserving the structure of draft assets and spatial layout of the canvas~(Sec.~\ref{sec:step4}).
A high-level illustration is shown in Fig.~\ref{fig:main}.

\subsection{Assets Generation}
\label{sec:step1}
Given a textual description $T$, we first employ a Large Language Model~(LLM) to parse the scene description and extract a set of objects $\mathcal{O}=\{o_1,o_2,...,o_n\}$. Each object corresponds to an independently drawable visual entity (such as stones, a lighthouse, etc.), while abstract or environmental elements~(e.g., mist) are excluded since they cannot be naturally and visibly represented as sketch assets.
To support subsequent draft generation and layout arrangement, each object is formulated as $o_i=(\mathrm{id}_i,d_i,z_i)$, where $\mathrm{id}_i$ denotes a unique object identifier, $d_i$ is a simple description sentence of the object, and  $z_i$ is its relative depth order. 

After obtaining the object set $\mathcal{O}$, we further generate a simple draft for each object.
Specifically, we employ SketchAgent~\cite{vinker2025sketchagent} as the draft generator. 
Given the description $d_i$, we predict an ordered sequence of sampled points on a canvas to describe the object outline. 
Then, by solving a least-squares optimization problem, these sampled points are fitted to a vector draft composed of cubic B\'ezier curves, producing object draft $g_i$.
Each draft is synthesized on a separate canvas and added to its corresponding object $o_i=(\mathrm{id}_i,d_i,z_i,g_i)$.
To prevent interference from other objects, we further inject a vectorized white background mask derived from the object's contour, enabling subsequent reasoning to assemble the scene in a collage-like manner.

\subsection{Operating Action Library}
\label{sec:step2}
To support interpretable and executable scene composition, we define a compact 
set of object-level operation codes that serve as the action space of the reasoning agent.
Given the current scene state, the agent produces a sequence of executable actions
to progressively arrange and revise the retrieved sketch assets.
An action is selected from six primitive operations:

\noindent
\texttt{\textbf{move}}$(o,x,y)$ moves object $o$ to canvas coordinate $(x,y)$.
     
\noindent
\texttt{\textbf{scale}}$(o,s)$ rescales object $o$ by a relative factor $s$.
    
\noindent
\texttt{\textbf{rotate}}$(o,\theta)$ rotates object $o$ by angle $\theta$ around its center.

\noindent
\texttt{\textbf{layer}}$(o,l)$ changes the depth order of object $o$ by assigning it to layer $l$.

\noindent
\texttt{\textbf{delete}}$(o)$ removes object $o$ from the scene composition.

\noindent
\texttt{\textbf{modify}}$(o,d)$ regenerates object $o$ based on description $d$.

Note that a single layout revision may require composing multiple primitives.
For example, placing a large object behind another may involve a
\texttt{scale}, followed by a \texttt{move} and a \texttt{layer} operation.
By keeping these primitives minimal and composable, the agent can express diverse layout manipulations with a compact action vocabulary, while each individual operation remains explicit and easy to verify.

\subsection{Agentic Layout Reasoning}
\label{sec:step3}
After generating draft assets, directly assembling drafts often produces implausible spatial relationships (\emph{e.g.}, inappropriate scales, incorrect occlusions). Therefore, \textit{SketchDirector} 
formulates scene sketching as an agentic layout reasoning process that employs a closed-loop \emph{Plan--Execute--Verify--Correct} process to iteratively improve scene composition. At each iteration, the
agent plans object-level programs, executes them, evaluates the resulting layout, and revises its
decisions when necessary, as illustrated in Figs.~\ref{fig:pipeline}.

\subsubsection{Planning and Execution. }
Planning aims to translate high-level layout intentions into an executable program.
We encode layout decisions as a sequence of functional tools rather than directly predicting object attributes, including \texttt{move()}, \texttt{scale()}, \texttt{rotate()}, \texttt{layer()}, \texttt{delete()} and \texttt{modify()}, enabling flexible and interpretable arrangement. 
To facilitate reliable spatial reasoning, the canvas is discretized into an $R\times R$ regular grid like previous works~\cite{sun2025layoutvlm, vinker2025sketchagent}. 
The agent infers the grid location of each object, which is deterministically mapped back to the corresponding canvas position during execution.
The agent generates programs by jointly reasoning over geometric and visual clues, where object attributes, grid positions, and bounding boxes provide spatial constraints, while rendered canvases and layout histories provide global context.
The generated operations are sequentially executed to update object attributes and scene states, producing an explicit and traceable composition trajectory, as illustrated in Fig.~\ref{fig:reason}.
This traceability allows the agent to identify problematic operations during subsequent verification and correction.

\subsubsection{Multimodal Verification.}
To ensure the layout from the agent meets spatial constraints and textual descriptions, we propose a hybrid verification with geometric and visual-semantic check. 
Geometric verification checks basic spatial validity by ensuring that all draft assets lie within the canvas boundaries and their sizes avoid extreme scaling anomalies. Visual-semantic verification complements geometric checks by assessing high-level semantic consistency. 
The details are described below.

\textbf{Geometric Verification.}
We evaluate the geometric plausibility of the generated layout from two aspects: boundary constraints and object scales.
Specifically, after executing the predicted layout program, each object is represented by its transformed bounding box on the canvas.
For boundary verification, we check whether the bounding box of each object is fully contained within the canvas region.
Any object exceeding the canvas boundary is recorded as a boundary violation with a violation count $I_{\mathrm{bound}}$.
The boundary verification is considered successful only when no violation is detected:

\[
P_{\mathrm{bound}}=\mathrm{True}
\quad \text{if}\quad
I_{\mathrm{bound}}=0 .
\]

For size verification, we examine the scale factor of each object to avoid extreme scaling artifacts.
Objects with scale factors outside the predefined range $[0.1,1.5]$ are recorded as size violations $I_{\mathrm{size}}$.
The size verification passes when:

\[
P_{\mathrm{size}}=\mathrm{True}
\quad \text{if}\quad
I_{\mathrm{size}}=0 .
\]

The geometric verification is satisfied only when both boundary and size constraints are fulfilled.

\textbf{Visual-Semantic Verification.}
We employ a Vision-Language Model (VLM) critic with a dual-image input consisting of the rendered scene canvas and the object collage.
The VLM produces scores $S_i\in[0,1]$ from five perspectives:
(1) $S_{\mathrm{sem}}$: semantic alignment between the generated scene and the target concept, evaluating whether the main objects and their intended meanings are correctly represented;
(2) $S_{\mathrm{comp}}$: spatial composition consistency, evaluating whether object placement, spacing, overlap, depth ordering, and overall arrangement form a coherent scene;
(3) $S_{\mathrm{compl}}$: object completeness, evaluating whether all required objects are present and recognizable according to the provided object collage;
(4) $S_{\mathrm{size}}$: size consistency, evaluating whether object scales and relative proportions are reasonable and physically plausible within the scene;
(5) $S_{\mathrm{aes}}$: visual aesthetics, evaluating the overall visual quality, including composition harmony, clarity, and visual appeal.
The overall evaluation score is computed as a weighted aggregation:
\begin{equation}
S_{\mathrm{visual}}
=
\sum \lambda_i S_i,
\quad
\sum\lambda_i=1,
\end{equation}
where $i\in\{\mathrm{sem},\mathrm{comp},\mathrm{compl},\mathrm{size},\mathrm{aes}\}$. 

The semantic verification is considered successful when:

\[
P_{\mathrm{visual}}=\mathrm{True}
\quad\text{if}\quad
S_{\mathrm{visual}}\geq\tau_{\mathrm{visual}}.
\]

Finally, the overall verification decision is determined by combining geometric and semantic assessments:

\begin{equation}
P_{\mathrm{overall}}
=
P_{\mathrm{bound}}
\land
P_{\mathrm{size}}
\land
P_{\mathrm{visual}} .
\end{equation}

Only layouts satisfying all verification criteria are accepted.
Otherwise, the detected violations and verification feedback are recorded and sent to the self-evolving correction module.
the agent then uses these failure signals together with previous execution trajectories to generate a revised layout program in the next planning iteration.

\subsubsection{Self-Evolving Correction.}
Different from contrastive experience for object-level sketch generation~\cite{xiao20263drawagent}, inspired by human learning, we propose a hierarchical self-evolving mechanism that Short-Term Failure Memory converts failure trajectories into corrective feedback and Long-Term Experience Memory transfers successful cross-scene trajectories into reusable experiences, which are incrementally stored and retrieved to improve subsequent layout planning and program generation.

\textbf{The Short-Term Failure Memory.}
The Short-Term Failure Memory maintains a trajectory-level record for each failed attempt. Each entry contains the generated program, intermediate rendering results, object-level attributes, verification feedback, and the corresponding failure category.
These records are discarded after successful correction and are only used within the current scene generation process.

\textbf{The Long-Term Experience Memory.}
The Long-Term Experience Memory stores successful layout experiences indexed by the corresponding object set.
For a new scene, relevant experiences are retrieved by measuring the Jaccard  similarity between the current object set and historical object sets~\cite{park2023generative,shinn2023reflexion}:

\begin{equation}
\mathrm{Sim}(\mathcal{S}_{\mathrm{new}},\mathcal{S}_{\mathrm{mem}})
=
\frac{
|\mathcal{S}_{\mathrm{new}}\cap\mathcal{S}_{\mathrm{mem}}|
}{
|\mathcal{S}_{\mathrm{new}}\cup\mathcal{S}_{\mathrm{mem}}|
}.
\end{equation}

For example, suppose the current scene contains
$\mathcal{S}_{\mathrm{new}}=\{\texttt{tree},\texttt{squirrel},\texttt{bird}\}$,
while a stored experience is indexed by
$\mathcal{S}_{\mathrm{mem}}=\{\texttt{tree},\texttt{squirrel},\texttt{monkey}\}$.
Their intersection is
$\{\texttt{tree},\texttt{squirrel}\}$ with cardinality $2$,
and their union is
$\{\texttt{tree},\texttt{squirrel},\texttt{bird},\texttt{monkey}\}$ with cardinality $4$.
Therefore,$\mathrm{Sim}=\frac{2}{4}=0.50$
Since the similarity reaches the predefined threshold
$\tau_{\mathrm{sim}}$,
this experience is retrieved and provided to the agent as a reference for subsequent layout planning.

\subsection{Detail Refinement}
\label{sec:step4}
After reasoning, the generated sketches achieve the desired spatial arrangement but contain coarse details.
To improve quality while preserving object structure and layout, we introduce a detail refinement using curve contour resampling to convert object contours into Bézier primitives and progressive joint optimization to refine local objects and the global canvas, as shown in Fig.~\ref{fig:main}.

\subsubsection{Curve Contour Resampling.}
To obtain a fixed number of optimization primitives, we resample object contours into $N_{\mathrm{target}}$ cubic B\'ezier segments. Unlike point-based B\'ezier resampling method~\cite{vinker2022clipasso}, our approach directly subdivides the existing vector contours without reconstructing their geometry.
Specifically, we convert foreground SVG paths into cubic Bézier curves, excluding background masks and degenerate segments. Given valid paths containing $N_{\mathrm{orig}}$ curves in total, we assume $N_{\mathrm{target}}\geq N_{\mathrm{orig}}$ and assign at least one segment to each original curve. The remaining budget is proportionally allocated first across paths and then across curves according to their arc lengths, using the largest-remainder method to ensure exactly $N_{\mathrm{target}}$ segments.
Each original curve is subdivided into its allocated number of cubic Bézier segments using the de Casteljau algorithm, which preserves the original contour geometry. The resulting ordered sequence of primitives is defined as

\begin{equation}
\begin{aligned}
\mathcal{B}
&=
\bigoplus_{k=1}^{K}
\bigoplus_{j=1}^{n_k}
\left(
B_{k,j}^{(1)},\ldots,B_{k,j}^{(N_{k,j})}
\right),\\
|\mathcal{B}|
&=
\sum_{k=1}^{K}\sum_{j=1}^{n_k}N_{k,j}
=N_{\mathrm{target}}.
\end{aligned}
\end{equation}

where \(\bigoplus\) denotes ordered concatenation, preserving the original path and curve ordering.
During subsequent refinement, the number and ordering of primitives remain fixed, while their control points are independently optimized. It provides a faithful starting point that retains object shapes and structural details, as illustrated in Fig.~\ref{fig:resampling}.

\subsubsection{Progressive Joint Optimization. }
Based on the resampled representation, each object draft is expressed as a fixed-topology set of B\'ezier primitives parameterized by control points $\theta_i$ and rendered as $I_i=R(\theta_i)$, where $R(\cdot)$ denotes the differentiable vector renderer~\cite{li2020differentiable}.
During optimization, only the control-point geometry is updated, while the curve ordering inherited from the draft remains fixed.

During local optimization, we apply Rectified-Flow Distillation Sampling~(RFDS~\cite{yang2025text}) for supervision:
\begin{equation}
\mathcal{L}^{i}_{obj}
=
\mathcal{L}_{RFDS}(I_{obj}^i,T^i_{\mathrm{obj}}),
\end{equation}
where $T^i_{\mathrm{obj}}$ corresponds to $d_i$, is the description of the object.

During global optimization, each object is incrementally composed with the canvas $I^{i}_{scene}=\mathcal{C}(I^{i-1}_{scene}, I_i,\ell_i)$ using layout attributes $\ell_i$, where $\mathcal{C}(\cdot)$ denotes differentiable composition. The scene is optimized with a global RFDS loss:
\begin{equation}
\mathcal{L}^{i}_{\mathrm{scene}}
=
\mathcal{L}_{\mathrm{RFDS}}
\left(
I^{i}_{\mathrm{scene}},
T^{i}_{\mathrm{acc}}
\right),
\end{equation}
where $T^{i}_{\mathrm{acc}}$ describes all objects incorporated into the accumulated canvas and their semantic and spatial relationships at the $i$-th optimization stage. Consequently, the text condition is progressively expanded as new objects are optimized.

At optimization round for each object, only the control points \(\theta_i\) of the current object are optimized.
The objective combines both object and global  supervision:
\begin{equation}
\mathcal{L}^{i}
=
\lambda_{obj}\mathcal{L}^{i}_{obj}
+
\lambda_{scene}\mathcal{L}^{i}_{scene},
\end{equation}
where $\lambda_{obj} $ and $\lambda_{scene} $ balance object-level semantic refinement and scene-level structural alignment.

\begin{figure*}[t]
    \centering
\includegraphics[width=1.0\linewidth]{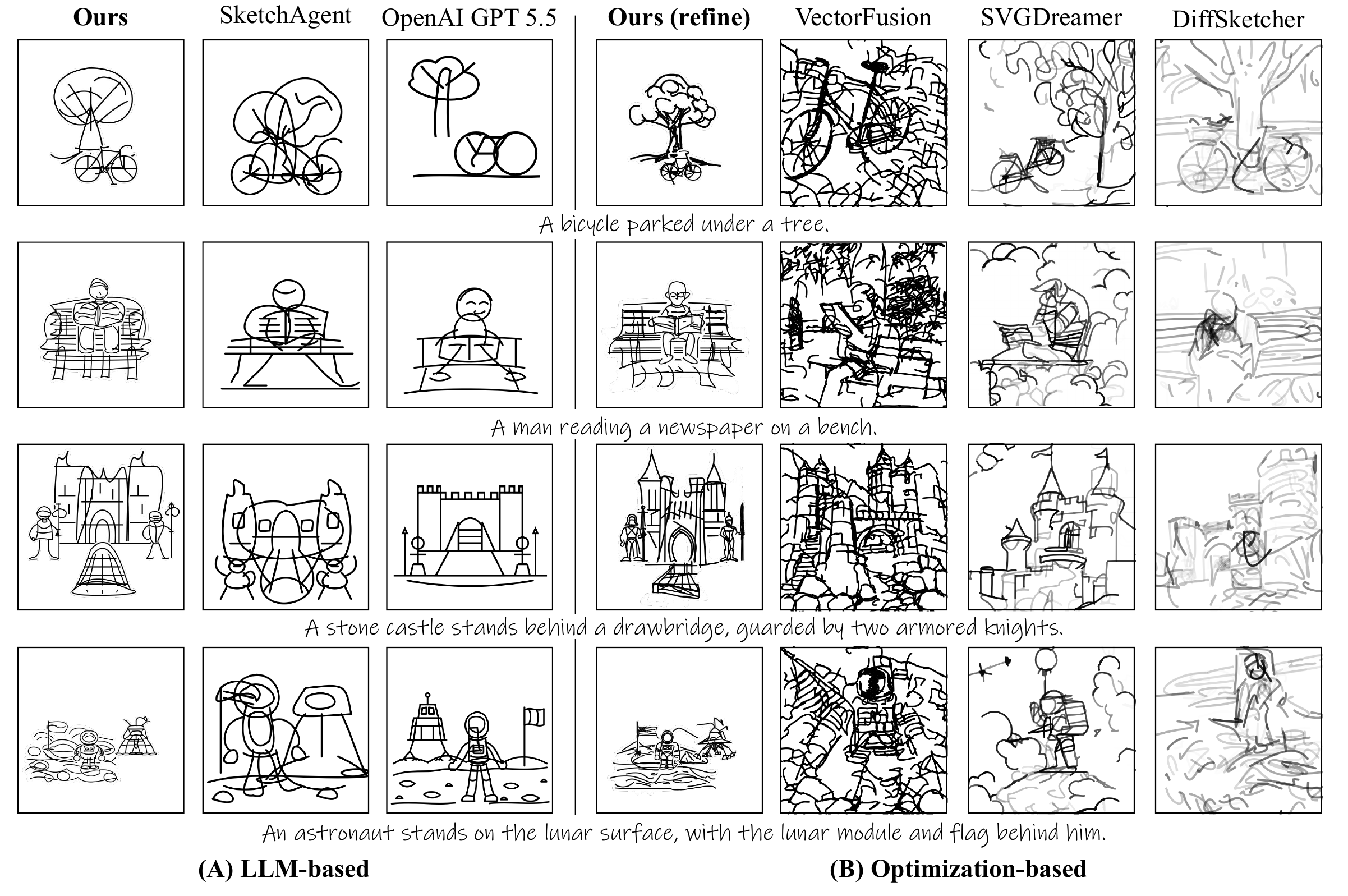} % Reduce the figure size so that it is slightly narrower than the column.
\vspace{-2em}
    \caption{\textbf{Qualitative comparison with existing methods.}
    Compared with \emph{LLM-based methods}, our sketch generation produces scene sketches with more explicit object organization, coherent layouts, and better visual-semantic alignment. 
    Moreover, compared with \emph{optimization-based methods}, our refinement improves sketch details while faithfully preserving the object layouts and spatial relationships in the draft.}
    \vspace{-1em}
    \label{fig:baseline_comparison}
\end{figure*}
\begin{figure*}[t]
    \centering
\includegraphics[width=1.0\linewidth]{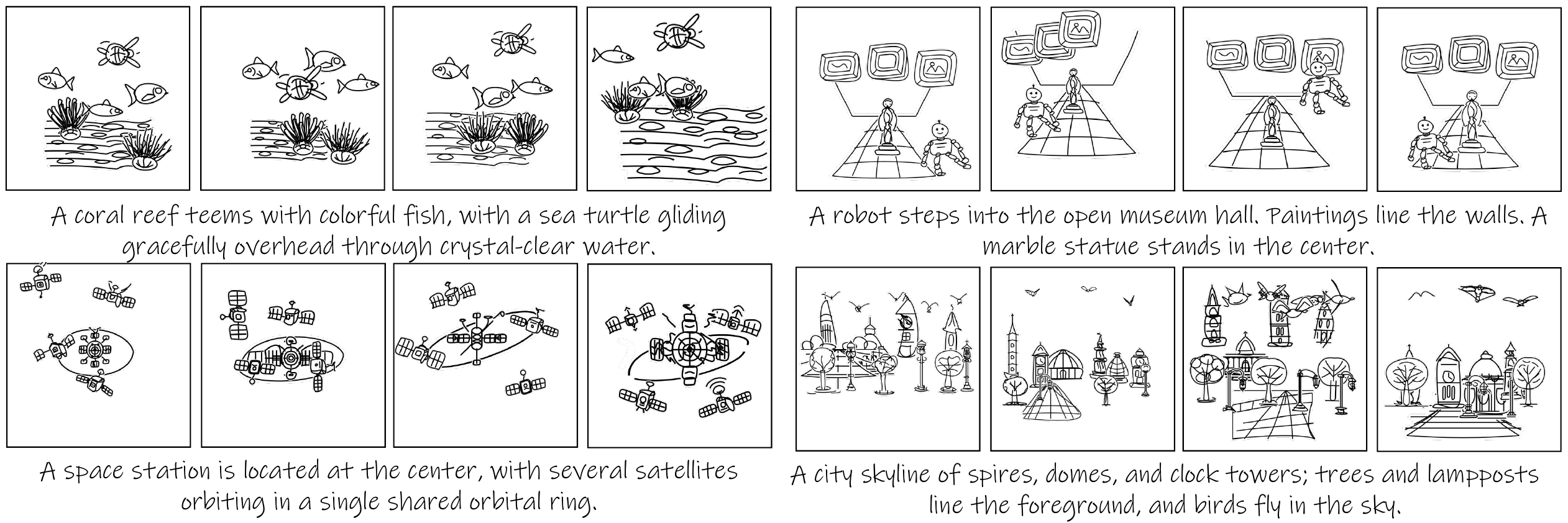} % Reduce the figure size so that it is slightly narrower than the column.
\vspace{-2em}
    \caption{\textbf{Scene sketch diversity generated by SketchDirector.}
    Our object-level reasoning enables flexible scene layouts from textual descriptions, by producing diverse spatial arrangements (the top row) and object configurations under similar semantic descriptions (the bottom row).}
    \vspace{-1em}
    \label{fig:layout_compare}
\end{figure*}

\section{Experiment}
\subsection{Experimental Setup}

\subsubsection{Implementation Details.} 
We deploy \textit{SketchDirector} on an NVIDIA GeForce RTX 4090 GPU.
We employ OpenAI GPT-5.5~\cite{openai2026gpt55} as the default agent and pydiffvg~\cite{li2020differentiable} as the differentiable vector renderer. 
All stroke primitives are parameterized as cubic B\'ezier curves defined by $4$ control points. 
The canvas is discretely configured as a $64 \times 64$ grid with each cell spanning $12 \times 12$ pixels, resulting in an overall rendered resolution of $780 \times 780$ pixels.
In terms of computational cost, asset generation takes approximately $20$ seconds per object, while planning, execution, and verification require approximately $30$, $12$, and $40$ seconds per layout attempt, respectively. Detail refinement takes approximately $5$ minutes per object with $400$ optimization iterations. Although this incurs additional computational overhead, the resulting independent SVG assets and executable layouts enable flexible and efficient object-level editing without full-scene regeneration.

\subsubsection{Benchmark.}
To evaluate the performance of scene sketch generation, we sample different multi-concept text as prompts spanning diverse themes and categories from \textit{MCSketch8K}, to comprehensively assess sketch quality.

\subsubsection{Baselines.}
We compare \textit{SketchDirector} against five baselines for scene sketch generation, categorized into LLM-based and optimization-based methods. 
\emph{LLM-based methods} include SketchAgent~\cite{vinker2025sketchagent} and OpenAI GPT-5.5~\cite{openai2026gpt55}; 
\emph{Optimization-based methods} include SVGDreamer~\cite{xing2024svgdreamer}, VectorFusion~\cite{jain2023vectorfusion} and DiffSketcher~\cite{xing2023diffsketcher}.

\subsubsection{Metrics.}
To assess the semantic and compositional quality of scene sketches, we adopt a multi-VLMs assessment to reduce evaluator bias.
We employ Qwen3.7-Plus~\cite{qwen2026qwen37plus}, Claude-Opus-4-7~\cite{anthropic2026claude47}, Gemini3.1-Pro~\cite{google2026gemini31pro}, and OpenAI GPT-5.5~\cite{openai2026gpt55} as judges.
Each VLM independently evaluates five dimensions:
Semantic Alignment~(Sem.), Spatial Composition~(Compo.), Object Completeness~(Compl.), Visual Aesthetics~(Aes.), and Size Consistency~(Size).
Scores are averaged across the four VLMs and all test samples for quantitative comparison.

\subsection{Experiment Results}
% \begin{table}[t]
%     \setlength{\tabcolsep}{2.5pt}
%     \small
%     \centering
%     \begin{tabular}{l|ccccc}
%     \toprule
%     \textbf{Method} 
%     & \textbf{Sem.} $\uparrow$
%     & \textbf{Compo.} $\uparrow$
%     & \textbf{Compl.} $\uparrow$
%     & \textbf{Aes.} $\uparrow$
%     & \textbf{Size} $\uparrow$\\
%     \midrule
%     SketchAgent  
%     & 0.742 & 0.598 & 0.589 & 0.714 & 0.693\\

%     OpenAI GPT-5.5  
%     & 0.911 & 0.730 & 0.811 & 0.813 & 0.820\\
    
%     VectorFusion  
%     & 0.466 & 0.252 & 0.338 & 0.417 & 0.412 \\

%     SVGDreamer  
%     & 0.690 & 0.546 & 0.620 & 0.650 & 0.695\\

%     DiffSketcher  
%     & 0.567 & 0.374 & 0.381 & 0.567 & 0.537\\

%     \midrule

%     SketchDirector  
%     & 0.916 & 0.772 & 0.821 & 0.836 & 0.821\\

%     SketchDirector (refine) 
%     & \textbf{0.923} & \textbf{0.788} & \textbf{0.843} & \textbf{0.869} & \textbf{0.842}\\

%     \bottomrule
%     \end{tabular}
%     \caption{\textbf{Comparison with baseline methods on multi-concept scene sketching.}
%     \textit{SketchDirector} achieves the highest performance across all five metrics.}
%     \label{tab:baseline}
% \end{table}

\begin{table}[t]
    \centering
    \setlength{\tabcolsep}{2pt}
    \small
    \resizebox{\columnwidth}{!}{
    \begin{tabular}{l|cccccc}
    \toprule
    \textbf{Method}
    & \textbf{Sem.} $\uparrow$
    & \textbf{Compo.} $\uparrow$
    & \textbf{Compl.} $\uparrow$
    & \textbf{Aes.} $\uparrow$
    & \textbf{Size} $\uparrow$
    & \textbf{Time} $\downarrow$\\
    \midrule

    SketchAgent
    & 0.742 & 0.598 & 0.589 & 0.714 & 0.693 & $\sim$20s\\

    OpenAI GPT-5.5
    & 0.911 & 0.730 & 0.811 & 0.813 & 0.820 & $\sim$20s\\

    VectorFusion
    & 0.466 & 0.252 & 0.338 & 0.417 & 0.412 & $\sim$17min\\

    SVGDreamer
    & 0.690 & 0.546 & 0.620 & 0.650 & 0.695 & $\sim$24min\\

    DiffSketcher
    & 0.567 & 0.374 & 0.381 & 0.567 & 0.537 & $\sim$13min\\

    \midrule

    SketchDirector
    & 0.916 & 0.772 & 0.821 & 0.836 & 0.821 & $\sim$6min\\

    SketchDirector+Refine
    & \textbf{0.923}
    & \textbf{0.788}
    & \textbf{0.843}
    & \textbf{0.869}
    & \textbf{0.842}
    & $\sim$21min\\

    \bottomrule
    \end{tabular}
    }
    \caption{\textbf{Comparison with baseline methods on
    multi-concept scene sketching.}
    SketchDirector achieves the best performance across
    all five metrics.}
    \label{tab:baseline}
\end{table}

We evaluate \textit{SketchDirector} through multi-VLMs assessment and user study to comprehensively measure quantitative performance and human preferences.
we sample 30 multi-concept scene descriptions from \textit{MCSketch8K} for evaluation.
Tab.~\ref{tab:baseline} shows the results between \textit{SketchDirector} and five baseline methods based on VLM assessment.
\textit{SketchDirector} achieves the highest performance across all five metrics.
Compared with \emph{LLM-based methods}, \textit{SketchDirector} further improves performance, showing the effectiveness of agentic reasoning. 
Compared with \emph{optimization-based methods}, \textit{SketchDirector}, OpenAI GPT-5.5, and SketchAgent obtain higher scores due to the capabilities of LLMs, indicating the importance of semantic understanding and high-level scene reasoning. 
The refinement enhances most metrics, demonstrating the effectiveness of detail refinement.

\begin{figure}[t]
\centering
\includegraphics[width=1.0\linewidth]{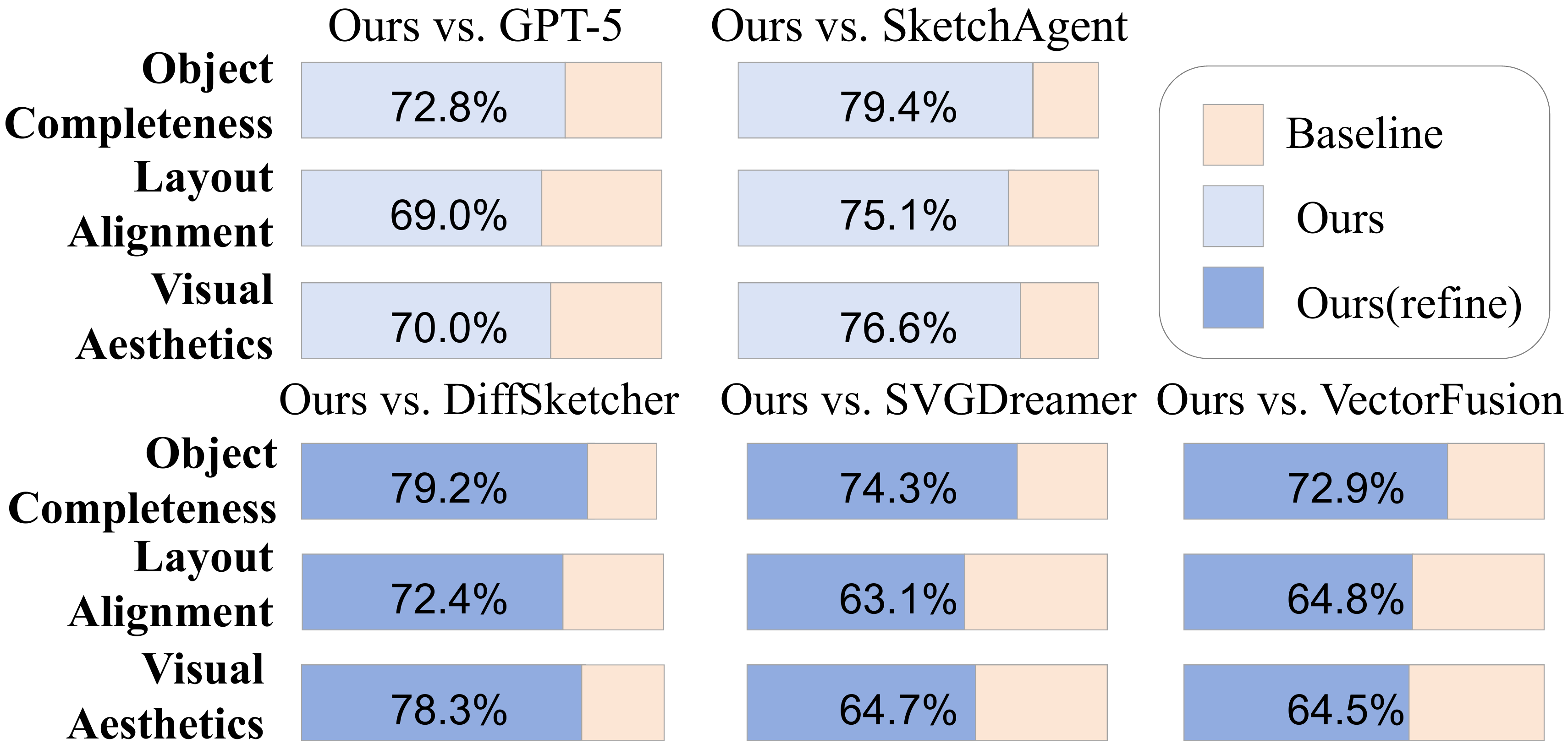} % Reduce the figure size so that it is slightly narrower than the column.
\vspace{-1em}
\caption{\textbf{User study results.} \textit{SketchDirector} preference over baseline methods across three criterias.}
\vspace{-1em}
\label{fig:userstudy}
\end{figure}

We also conduct a user study with 68 participants recruited from the university community.
We sample 15 diverse scene prompts from the \textit{MCSketch8K} and generate sketches using \textit{SketchDirector} and each baseline. 
For fair comparison, i.e., to isolate the effect of refinement, we consider two settings. 
Specifically, \textit{SketchDirector} without refinement is compared with \emph{LLM-based methods}, while \textit{SketchDirector} with structural refinement~(Ours~(refine)) is compared with \emph{optimization-based methods}. 
For each pair, participants viewed a description and two randomly ordered sketches and answered three questions:
(1) (Object Completeness) \textit{Which sketch better represents the objects in the text?}  (2) (Layout alignment) \textit{Which sketch has better layout alignment?} (3) (Visual Aesthetic) \textit{Which sketch looks more aesthetically pleasing?} 
Overall, we collect 5100 responses per criterion. 
Fig.~\ref{fig:userstudy} reports \textit{SketchDirector} preference rates for the three questions.
Across both settings and all baselines, participants consistently favored our results.

Fig.~\ref{fig:baseline_comparison} shows the qualitative comparison with baseline methods. 
\emph{LLM-based methods} show strong semantic understanding and object-level depiction, but may struggle with object disentanglement and visual quality.
SketchAgent~\cite{vinker2025sketchagent} generates clean sketches through sequential drawing actions, but often causes object overlap in complex multi-concept scenes, while OpenAI GPT-5.5~\cite{openai2026gpt55} generates recognizable structures with rigid line styles. 
In contrast, \textit{SketchDirector} generates sketches with clearer object boundaries and better visual presentation.
\emph{Optimization-based methods} often generate sketches with fragmented strokes and ambiguous structures, due to the lack of explicit object-level representations. 
VectorFusion~\cite{jain2023vectorfusion} and SVGDreamer~\cite{xing2024svgdreamer} tend to produce dense and irregular paths, while DiffSketcher~\cite{xing2023diffsketcher} captures coarse semantics but often results in incomplete structures or limited details. 
Moreover, Fig.~\ref{fig:layout_compare} shows that \textit{SketchDirector} can generate diverse scene layouts from the same prompt.

\subsection{Ablation Studies. }
To systematically evaluate the individual contributions of key components in \textit{SketchDirector}, we conduct ablation studies on another 30 multi-concept scene prompts from \textit{MCSketch8K}. 
Specifically, we investigate the impact of different VLM backbones on asset generation and layout reasoning, the contributions of individual components within the agentic reasoning loop, and the effectiveness of contour resampling and progressive joint optimization in detail refinement.
% We investigate the effects of Agentic Layout Reasoning, detail Refinement, and VLM backbones, with OpenAI GPT-5.5~\cite{openai2026gpt55} as the default backbone for pairwise A/B testing and metric evaluation.

\subsubsection{Impact of VLM Backbones.}
\begin{figure}[t]
    \centering
\includegraphics[width=\linewidth]{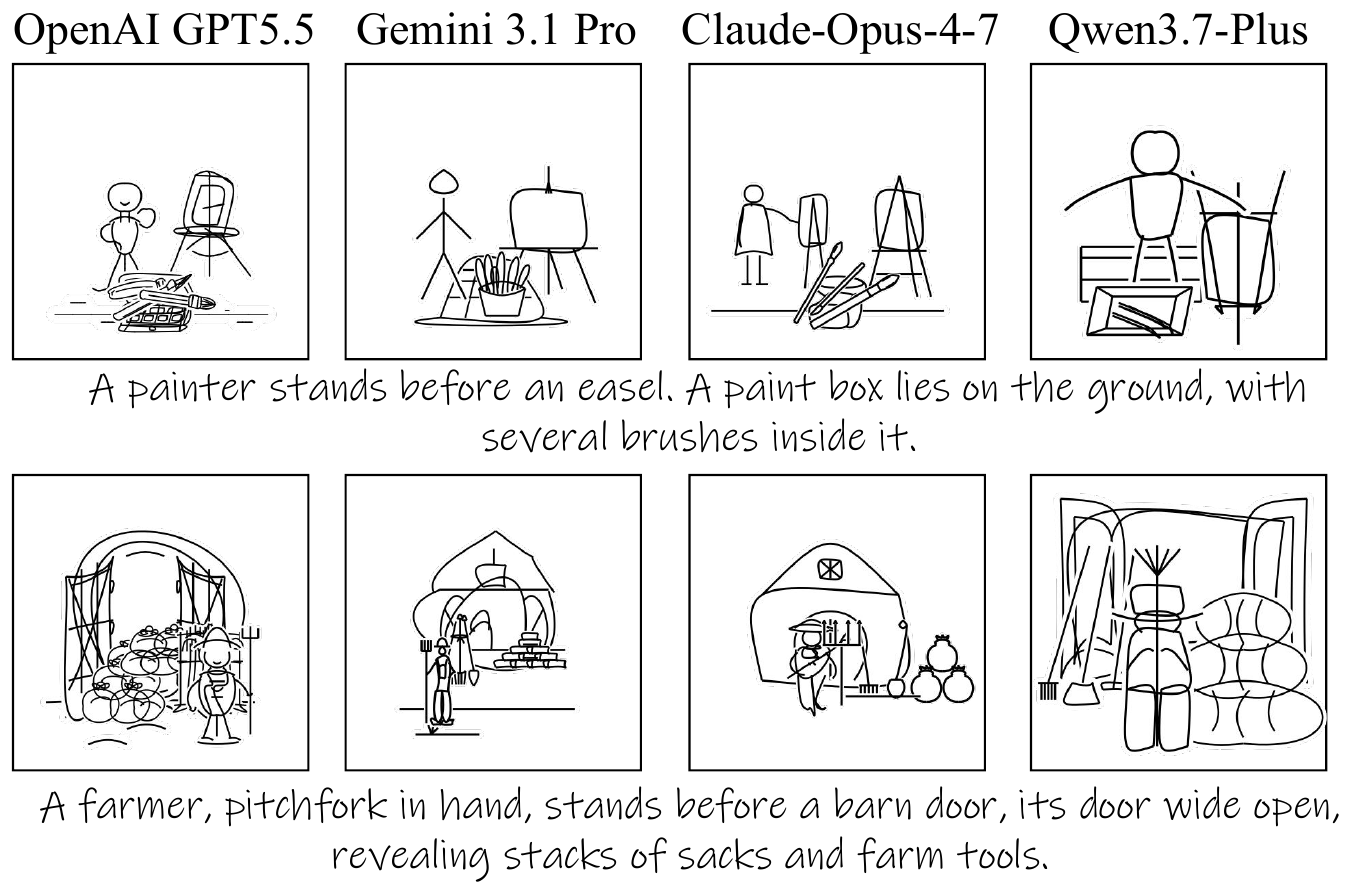} % Reduce the figure size so that it is slightly narrower than the column.
    \caption{\textbf{Qualitative comparison of VLM backbones.}
    Visual results highlight differences in object-level fidelity, scene composition, and visual details across backbone models.}
    \label{fig:backbone}
\end{figure}

To evaluate the generalizability and robustness of \textit{SketchDirector} across different foundational models,we evaluate four state-of-the-art backbones: OpenAI GPT-5.5~\cite{openai2026gpt55}, Gemini-3.1-pro~\cite{google2026gemini31pro}, Claude-Opus-4-7~\cite{anthropic2026claude47}, and Qwen3.7-Plus~\cite{qwen2026qwen37plus}. 
Figure~\ref{fig:backbone} and Table~\ref{tab:backbone} show that that different backbones exhibit varying capabilities in spatial organization, visual-semantic alignment, and reasoning efficiency. 
OpenAI GPT-5.5 achieves the highest scores across all verification metrics and requires the fewest reasoning iterations, demonstrating stronger layout planning and scene composition capabilities. 
Gemini-3.1-Pro and Claude-Opus-4-7 achieve relatively strong semantic alignment but require more iterations to refine their layouts. Their generated sketches generally capture the main scene concepts, yet exhibit fewer visual details . In contrast, Qwen3.7-Plus obtains lower verification scores and requires substantially more reasoning iterations.
Overall, these results demonstrate that \textit{SketchDirector} can operate with different backbones, while stronger reasoning capabilities contribute to more accurate scene composition and efficient convergence.
\begin{table}[t]
    \setlength{\tabcolsep}{2.5pt}
    \small
    \centering
    \resizebox{\columnwidth}{!}{
    \begin{tabular}{l|ccccccc}
    \toprule
    \textbf{Method} 
    & \textbf{Sem.}$\uparrow$
    & \textbf{Comp.}$\uparrow$
    & \textbf{Compl.}$\uparrow$
    & \textbf{Aes.}$\uparrow$
    & \textbf{Size}$\uparrow$
    & \textbf{Iter.}$\downarrow$
    & \textbf{Ver.}$\uparrow$\\
    \midrule

    Gemini-3.1-pro 
    & 0.858 & 0.723 & 0.723 & 0.819 & 0.707  & 5.332 & 0.535 \\

    Claude-Opus-4-7   
    & 0.879 & 0.681 & 0.791 & 0.739 & 0.721 & 5.843 & 0.477 \\

    Qwen3.7-Plus
    & 0.629 & 0.477 & 0.526 & 0.552 & 0.507 & 10.743 & 0.436 \\

    OpenAI GPT-5.5 
    & \textbf{0.926} & \textbf{0.795} & \textbf{0.811} & \textbf{0.871} & \textbf{0.785} & \textbf{1.656} & \textbf{0.803} \\

    \bottomrule
    \end{tabular}
    }
    \vspace{-1em}
    \caption{\textbf{Quantitative comparison of VLM backbones.}
    Scores and reasoning iterations evaluate the layout reasoning quality and efficiency across backbone models. }
    \label{tab:backbone}
\end{table}
\begin{figure}[t]
\centering
\includegraphics[width=1.0\linewidth]{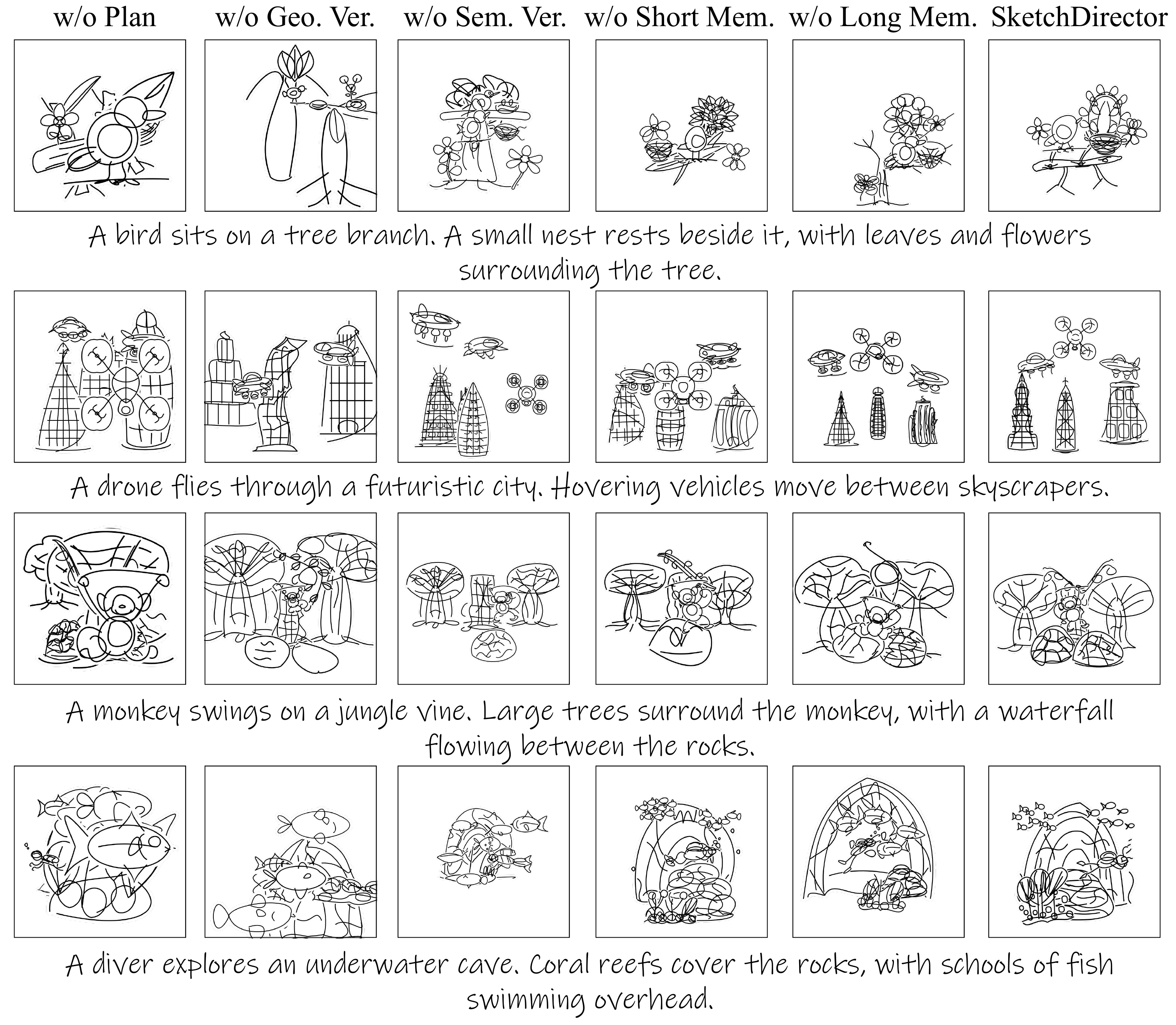} % Reduce the figure size so that it is slightly narrower than the column.
\vspace{-1em}
\caption{\textbf{Qualitative ablation of agentic layout reasoning.}
Visual results illustrate how individual components affect scene composition and object-level spatial coherence.}
\vspace{-1em}
\label{fig:abla_component}
\end{figure}
\begin{table}[t]
    \setlength{\tabcolsep}{2.5pt}
    \small
    \centering
    \resizebox{\columnwidth}{!}{
    \begin{tabular}{l|ccccccc}
    \toprule
    \textbf{Method} 
    & \textbf{Sem.}$\uparrow$
    & \textbf{Comp.}$\uparrow$
    & \textbf{Compl.}$\uparrow$
    & \textbf{Aes.}$\uparrow$
    & \textbf{Size}$\uparrow$
    & \textbf{Iter.}$\downarrow$
    & \textbf{Ver.}$\uparrow$\\
    \midrule

    w/o Plan  
    & 0.676 & 0.503 & 0.571 & 0.65 & 0.468 & 12.375 & 0.357 \\

    w/o Geo. Ver.  
    & 0.698 & 0.550 & 0.570 & 0.660 & 0.589 & 1.857 & 0.633 \\

    w/o Sem. Ver.   
    & 0.603 & 0.512 & 0.645 & 0.558 & 0.547 & 1.438 & 0.468 \\

    w/o Short Mem.  
    & 0.737 & 0.658 & 0.629 & 0.654 & 0.681 & 2.875 & 0.635 \\

    w/o Long Mem. 
    & 0.781 & 0.679 & 0.697 & 0.762 & 0.729 & 1.600 & 0.650 \\

    \midrule

    SketchDirector  
    & \textbf{0.839} & \textbf{0.765} & \textbf{0.732} & \textbf{0.768} & \textbf{0.742} & \textbf{1.375} & \textbf{0.873} \\

    \bottomrule
    \end{tabular}
    }
    \vspace{-1em}
    \caption{\textbf{Quantitative ablation of agentic layout reasoning.}
    Performance comparisons across metrics and reasoning iterations demonstrate the contributions of individual components to layout quality and reasoning efficiency.}
    \vspace{-1em}
    \label{tab:component}
\end{table}
\begin{figure}[t]
    \centering
\includegraphics[width=1.0\linewidth]{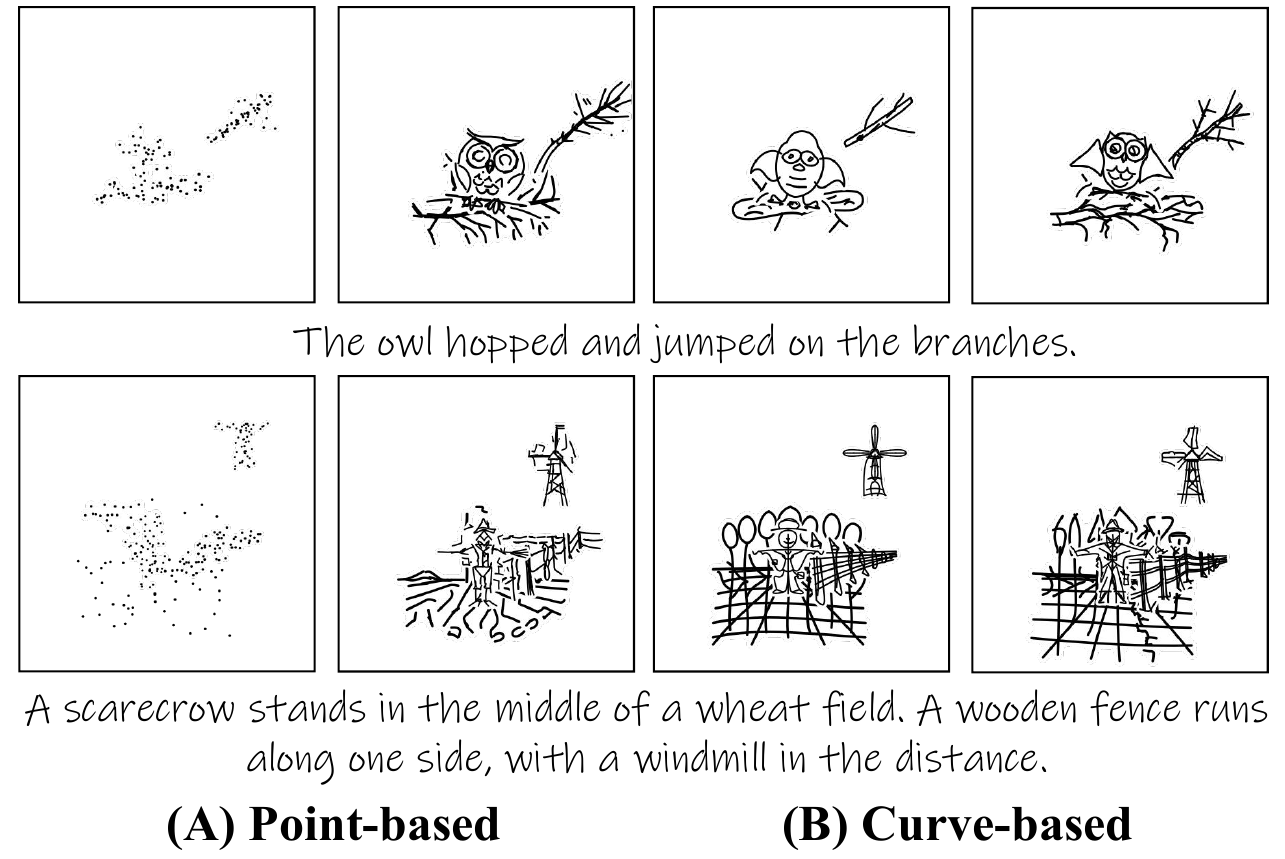} % Reduce the figure size so that it is slightly narrower than the column.
\vspace{-2em}
\caption{\textbf{Comparison of point-based and curve-based contour resampling method.} Curve-based contour resampling preserves object shapes and structural details, providing a more stable initialization for subsequent refinement.}
\label{fig:resampling}
\end{figure}
\subsubsection{Impact of Agentic Layout Reasoning.}
Figure~\ref{fig:abla_component} and Table~\ref{tab:component} show that removing any component leads to performance degradation, confirming that all modules in agentic reasoning are essential for high-quality scene sketch generation.
Removing planning  (\textit{w/o Plan})  leads to severe layout degradation.
Without executable scene programs, the model struggles to organize object attributes, resulting in frequent object overlaps, unreasonable spatial arrangements, and boundary violations.
These demonstrate that converting high-level semantic intentions into structured layout operations is essential for reliable scene composition.
Disabling geometric verification  (\textit{w/o Geo. Ver.}) mainly affects physical plausibility.
Although the generated objects remain semantically relevant, their positions and sizes may violate basic spatial constraints, such as objects extending beyond the canvas or unrealistic relative scales between objects.
This indicates that geometric constraints provide necessary low-level guidance for maintaining valid scene structures.
Without visual-semantic verification (\textit{w/o Sem. Ver.}), the agent lacks high-level visual feedback during layout refinement.
Consequently, generated scenes may contain incorrect object relationships, missing semantic interactions, or unnatural occlusion orders.
The comparison highlights the importance of multimodal evaluation in aligning generated layouts with the intended scene description.
Furthermore, memory modules improve the agent's ability to learn from previous reasoning trajectories.
Removing short-term memory (\textit{w/o Short Mem.}) does not cause significant visual layout degradation, but leads to repeated exploration of similar unsuccessful layouts, increasing the number of reflection iterations.
Removing long-term memory (\textit{w/o Long Mem.}) also preserves comparable visual structures in many cases, yet prevents the agent from retrieving reusable layout experiences from related scenes, resulting in less effective layout composition.
Overall, these qualitative examples verify that each component contributes complementary capabilities, enabling more reliable and controllable scene reasoning.

\subsubsection{Impact of Detail Refinement.}

\begin{figure}[t]
    \centering
\includegraphics[width=1.0\linewidth]{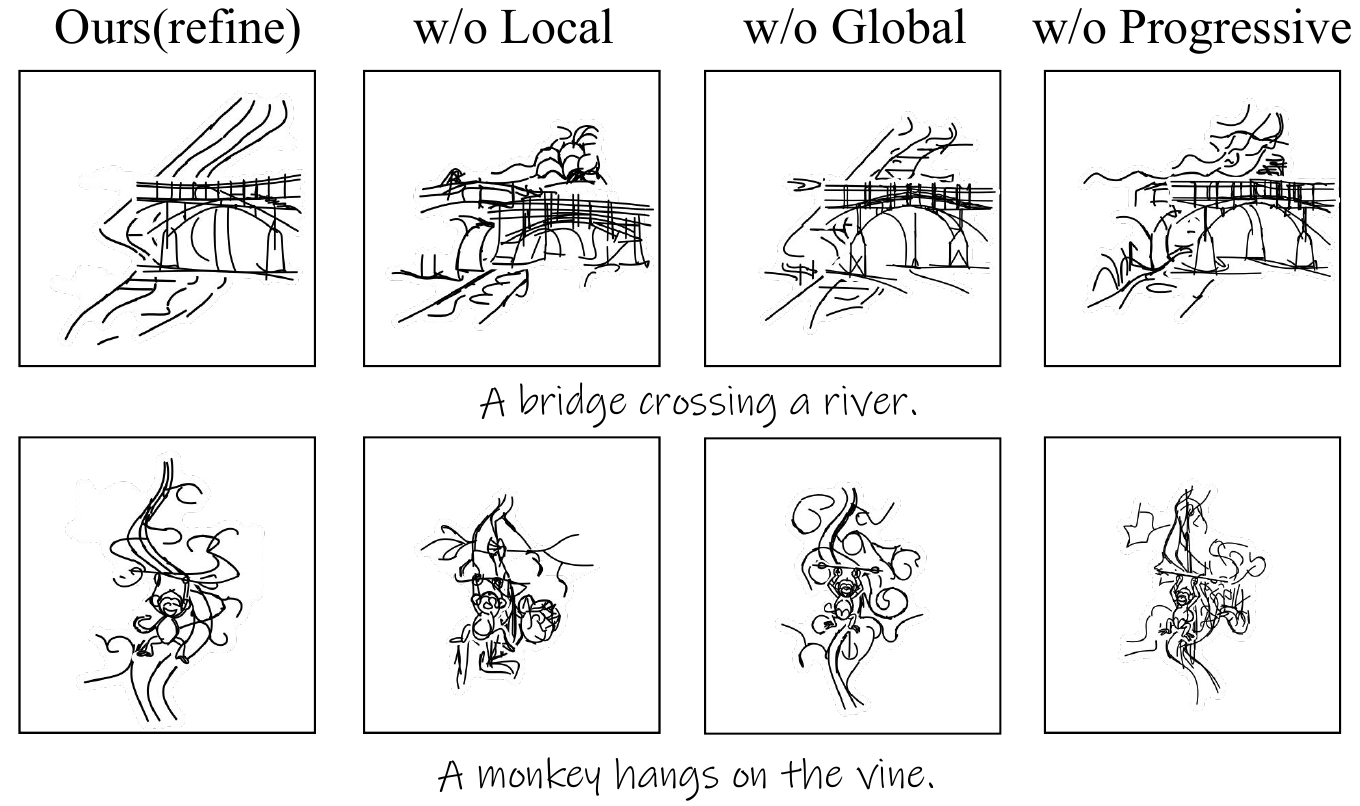} % Reduce the figure size so that it is slightly narrower than the column.\
\vspace{-2em}
\caption{\textbf{Qualitative ablation of progressive joint optimization.}
Removing local supervision, global supervision, or progressive optimization leads to degraded object fidelity and scene-level coherence.}
\vspace{-1em}
\label{fig:joint_optimize}
\end{figure}

Fig.~\ref{fig:resampling} provides additional comparisons between point-based and curve-based sampling.
Point-based sampling uniformly samples sparse control points from object regions without considering the underlying contours.
During optimization, these points may drift independently, causing geometric distortions and loss of characteristic object shapes.
In contrast, our curve contour resampling method derives B\'ezier primitives directly from object contours.
By preserving the original contour topology, the curves provide a more stable optimization space. 
Fig.~\ref{fig:joint_optimize} presents additional comparisons of different optimization strategies.
Removing local supervision (\textit{w/o Local}) optimizes objects only according to global objectives, allowing unrelated semantic information to interfere with individual object.
This often results in distorted local structures.
Removing global supervision (\textit{w/o Global}) focuses only on individual object while ignoring interactions among objects.
Although individual objects may remain recognizable, the resulting scenes exhibit weaker spatial coherence, such as unnatural object relationships.
Without progressive optimization (\textit{w/o Progressive}), local and global objectives are optimized simultaneously from the beginning.
The strong coupling between object-level refinement and scene-level consistency introduces optimization conflicts, causing unstable updates and reduced object fidelity.
By progressively incorporating objects, our method enables stable local refinement first and gradually improves global scene integration.

\subsection{Limitations.}
\begin{figure}[t]
\centering
\includegraphics[width=1.0\linewidth]{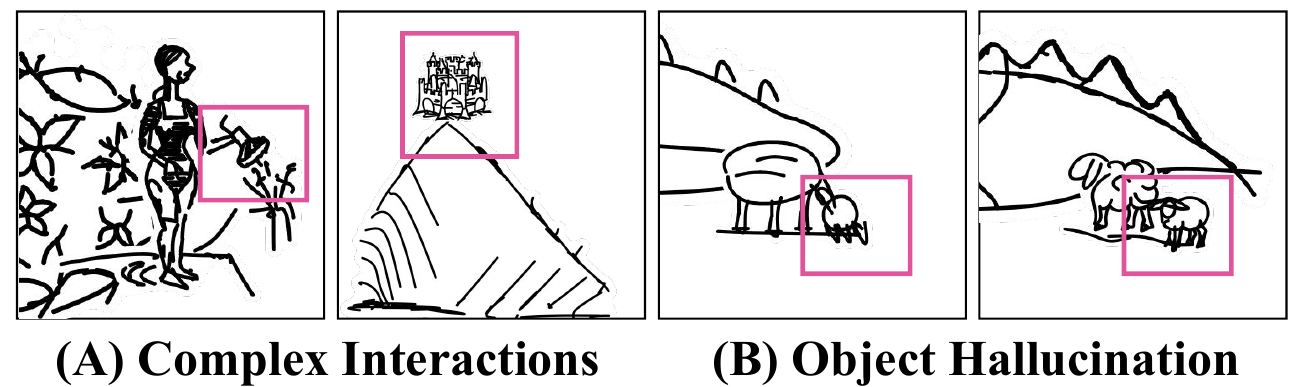} % Reduce the figure size so that it is slightly narrower than the column.
\vspace{-1.5em}
\caption{\textbf{Limitations of \textit{SketchDirector}.}
(A) Challenges in
modeling complex physical and semantic interactions. (B) Object hallucination during refinement.}
\label{fig:limitation}
\end{figure}

Although \textit{SketchDirector} generates multi-concept scene sketches, several limitations remain. First, it struggles with complex physical and semantic relations, such as interactions between a woman and a kettle or a castle located on a hill (Fig.~\ref{fig:limitation}~(A)). Stronger relational and interaction modeling may improve scene understanding. Second, refinement may hallucinate redundant instances, leading to object inconsistency, such as duplicating a sheep (Fig.~\ref{fig:limitation}~(B)). Instance-aware constraints could help preserve object identity and count. Third, multimodal verification and iterative optimization incur computational costs, particularly when multiple refinement rounds are required. Efficient verification and optimization strategies will be explored in future work.

\section{Conclusion}

We present \textit{SketchDirector}, an agentic framework that emulates the human sketching workflow. Designed for complex scene sketch generation, \textit{SketchDirector} converts textual descriptions into editable vector assets and executable layout operations. It consists of three stages: concept representation for generating object-level draft assets, agentic reasoning with iterative\emph{Plan--Execute--Verify--Correct} and self-evolving memory for reliable spatial composition, and detail refinement for progressively optimizing B\'ezier-based vectors while preserving the original sketch structure. We further introduce \emph{MCSketch8K}, a benchmark containing editable vector assets and composition annotations for multi-concept scene sketch generation. Extensive experiments demonstrate that \textit{SketchDirector} outperforms existing methods in semantic and compositional quality while effectively preserving structural integrity during refinement. 

\bibliography{Bibliography-File}

\end{document}